\documentclass[letterpaper, 10 pt, conference]{ieeeconf}  

\IEEEoverridecommandlockouts                              

\usepackage{graphicx} 
\usepackage{amsmath} 
\usepackage{amssymb}  
\usepackage{gensymb}
\usepackage{bm}
\usepackage{flushend}

\newcommand{\turbineparams}{\mathbf{\theta}}
\newcommand{\turbinecentre}{\mathbf{c}}
\newcommand{\turbineheight}{h}
\newcommand{\turbineheading}{\omega}
\newcommand{\turbinenacelle}{r}
\newcommand{\turbinebladelength}{b}
\newcommand{\turbinebladerot}{\phi}

\newcommand{\model}{\mathcal{M}}
\newcommand{\pointmodel}{\mathcal{P}}
\newcommand{\pointmodelpoint}{\mathbf{p}}
\newcommand{\linemodel}{\mathcal{L}}
\newcommand{\linemodelline}{\mathbf{l}}

\newcommand{\vertex}{\mathbf{v}}
\newcommand{\cammat}{\mathbf{K}}
\newcommand{\projimage}{\mathbf{P}}

\newcommand{\pointterm}{E^{\model}_{\text{unary (1)}}}
\newcommand{\imageterm}{E^{\model}_{\text{unary (2)}}}

\title{\LARGE \bf
	Simultaneous drone localisation and wind turbine model fitting\\ during autonomous surface inspection
}

\author{Oliver Moolan-Feroze$^{1}$, Konstantinos Karachalios$^{2}$, Dimitrios N. Nikolaidis$^{2}$, and Andrew Calway$^{1}$
\thanks{$^{1}$Oliver Moolan-Feroze and Andrew Calway are with the Department of Computer Science, University of Bristol, 75 Woodland Road, Bristol, BS8 1UB, United Kingdom
        {\tt\footnotesize oliver.moolan-feroze@bristol.ac.uk, andrew.calway@bristol.ac.uk}}%
	\thanks{$^{2}$Konstantinos Karachalios and Dimitrios N. Nikolaidis are with Perceptual Robotics, 5 Hope Road, Bristol, UK, BS3 3NZ, United Kingdom
        {\tt\footnotesize kostas@perceptual-robotics.com, dimitris@perceptual-robotics.com}}%
}

\begin{document}

\maketitle
\thispagestyle{empty}
\pagestyle{empty}

\begin{abstract}
	We present a method for simultaneous localisation and wind turbine model fitting for a drone performing an automated surface inspection. We use a skeletal parameterisation of the turbine that can be easily integrated into a non-linear least squares optimiser, combined with a pose graph representation of the drone's 3-D trajectory, allowing us to optimise both sets of parameters simultaneously. Given images from an onboard camera, we use a CNN to infer projections of the skeletal model, enabling correspondence constraints to be established through a cost function. This is then coupled with GPS/IMU measurements taken at key frames in the graph to allow successive optimisation as the drone navigates around the turbine. We present two variants of the cost function, one based on traditional 2D point correspondences and the other on direct image interpolation within the inferred projections. Results from experiments on simulated and real-world data show that simultaneous optimisation provides improvements to localisation over only optimising the pose and that combined use of both cost functions proves most effective.
\end{abstract}

\section{INTRODUCTION}\label{sec:introduction}

Over the operational lifetime of a wind turbine, it will likely incur a wide range of structural damage~\cite{Ribrant2007} due to adverse weather and events such a lightning strikes and bird collisions. For operational efficiency, this damage needs to be detected and fixed as soon as possible~\cite{GarciaMarquez2012}. This currently relies on manual inspections using climbing equipment, or the use of telephoto lenses from the ground. From the danger incurred during manual inspection, to the inability to get full coverage of the turbine surface using ground based cameras, it is evident that both methods have downsides. 

One way of addressing the problem is the use of unmanned autonomous vehicles (UAVs)  -- or drones -- to perform inspections~\cite{Wang2017}. This provides a number of benefits. It is safe, as no human personnel are required to climb the turbine. Also, as it is able to navigate freely around the turbine, it is able to obtain a complete view of the surface, including from above. Another significant benefit is the abilitiy to perform consistent and repeatable inspections. 

\begin{figure}[t]
	\centering
	\includegraphics[width=1.0\linewidth]{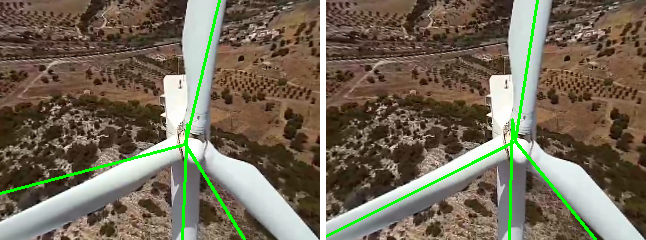}
	\caption{Figure showing the benefit of simultaneous model fitting and pose estimation. The green lines show the reprojection of the wind turbine using the optimised model parameters and the pose estimates. Left) Turbine reprojection when only the pose is optimised. Right) Turbine reprojection when both pose and model parameters are optimised.}
	\label{fig:model_image}
\end{figure}
Key to the drone's inspection performance is its ability to navigate around the turbine. To enable this, it needs an accurate estimate of its 3-D pose w.r.t the turbine, as well as an accurate estimate of the turbine's structure and configuration. Having these allows faster and safer navigation, more consistent inspections, and provides important information needed for post-processing operations such a image stitching and automatic default detection. Although the drone has access to a global positioning system (GPS) and inertial measurement units (IMUs) to aid in localisation, it has been shown that incorporating image measurements can improve accuracy significantly~\cite{Leutenegger2015,Moolan-Feroze2019}.

We present a method that is able to simultaneously optimise the drone's pose, as well as fit a model of the turbine during inspection. These two processes are complementary, in that the estimation of the turbine model is needed for accurate localisation, and localisation is required for fitting the model. The joint optimisation is achieved using a skeletal parameterisation of the turbine structure, accompanied with a set of functions that are able to instantiate the model. These functions, along with a pose graph representation of the drone's localisation are integrated into a non-linear least squares optimiser. We also present a novel cost function based on direct image interpolation that is better suited to this application than the typical cost based on 2D corresponding points. The work builds on our localisation method previously described in \cite{Moolan-Feroze2019}.

There are two main contributions. The first is the integration of the turbine model fitting process into the non-linear least squares optimiser. The skeletal representation and set of functions were carefully chosen so that they could be easily integrated into the optimiser as well as providing a model that fully describes the main structures of the turbine and is sufficiently general so as to be used with a range of turbine configurations and types. The second contribution is the use of an interpolation-based cost measure, which proves to be effective for optimising over line structures in the skeletal model and complements the point based approach described in \cite{Moolan-Feroze2019}.

In Section~\ref{sec:previous-work} we highlight some of the relevant literature related to this work. In Section~\ref{sec:method}, we give a detailed description of the proposed method. This includes the parameterisation of the model and accompanying functions in Section~\ref{sec:method-model}, the production of the image measurements using a CNN in Section~\ref{sec:method-projection}, and the joint optimisation of the pose and model parameters in Section~\ref{sec:method-optimisation}. In Section~\ref{sec:evaluation}, we show our evaluation of the method. This was done using both simulated data in Section~\ref{sec:evaluation-simulated} as well as real data in Section~\ref{sec:evaluation-real}. In Section~\ref{sec:conclusions} we present some conclusions as well as potential avenues for future work.

\section{PREVIOUS WORK}\label{sec:previous-work}

Likely due to the relative novelty of the application and the difficulty in acquiring data, there is little published research on the autonomous inspection of wind turbines using drones. Moreover, the works that tackle this problem have a number of important omissions. In~\cite{Stokkeland2015}, a method is presented for the prediction of the configuration of the wind turbine during an initial approach of the drone. They use a combination of the Hough Transform to extract edges and estimate the state of the turbine and a Kalman Filter to keep track of the drone's position. One problem with this work is that it only addresses the initial approach of the drone, and does not attempt to estimate the parameters during the remaining -- and most important -- part of the inspection. In~\cite{Schafer2016}, the authors suggest the use of LiDAR sensors to both estimate the state of the turbine and the position of the drone. The turbine model is represented as a voxel grid which is filled in using a Bayesian update scheme during the inspection process. The grid is then used for path planning. One drawback of this work is that it is only applied in simulation. Furthermore, the authors make no attempt at the simultaneous localisation of the drone. 

Although not directly related to wind turbine inspection, there are a number of methods that attempt simultaneous localisation and model fitting. These works are mainly focussed on the use of prior knowledge to improve the performance of simultaneous localisation and mapping (SLAM). In~\cite{Melbouci2016}, a method using RGBD data is proposed which makes use of a partial scene model that can constrain the optimiser, improving performance when using noisy and inaccurate depth data. Similarly, the work by Loesch et. al.~\cite{Loesch2018}, describes the integration of known objects into a SLAM process. Using an edglet-based feature description of the object, the method is able to constraint the SLAM process and reduce absolute drift in the localisation estimates.

Research into articulated object pose detection is relevant to the model fitting part of our method. Pauwels et. al.~\cite{Pauwels2014} propose an Iterative Closest Point (ICP) method that is tailored to articulated models. This is achieved using a set of constraints on the objects which restrict the possible movements between object parts based on the current view. Constraints are also used in~\cite{Michel2015}, in the form of a kinematic chain. This allows the authors to fully localise multi part objects by estimating only the parameters of the chain rather than the full 6DoF poses of the parts. To aid in this they make use of a random forest to predict object part probabilities and object coordinates. Katz et. al.~\cite{Katz2013} also make use of a kinematic chain, extracting and tracking visual features as a robot is interacting with articulated objects. Using clustering, tracked features are grouped into sets that represent different parts of the articulated object. From these, trajectories are established which define the types of constraints needed by different object parts. 

The work presented here is an extension of the localisation method described in~\cite{Moolan-Feroze2019}, which optimises the pose of the drone using a fixed skeletal turbine model. It used a convolutional neural network (CNN) to infer the projection of the model in images captured onboard, which are then integrated into a pose graph optimisation framework. We follow a similar approach in this work, except that we simultaneously estimate the turbine model parameters, producing improved localisation results and providing greater flexibility in the use of the method. In the following, we omit some of the detail regarding the CNN due to space constraints and refer readers to \cite{Moolan-Feroze2019} for a fuller description.

%
\section{METHOD}\label{sec:method}

The joint turbine model fitting and pose estimation process is performed throughout the duration of an inspection flight. We model the problem as a non-linear least squares optimisation over the set of model parameters and pose parameters. In Section~\ref{sec:method-model} we detail the parameterisation of the wind turbine model, and describe the functions that instantiate the model. In Section~\ref{sec:method-projection} we describe our CNN-based model projection estimation method. The projection estimates allow us to easily find correspondences between the model and image features. In Section~\ref{sec:method-optimisation} we describe the optimisation process, detailing the cost function and how it is optimised.

\subsection{Turbine Model Parameterisation}\label{sec:method-model}

There are two aims determining the selection of the parameterisation of the wind turbine model. The first is that it should be as minimal as possible to simplify the optimisation process. The second is that the parameterisation must be complete, in that it fully describes the structure of the turbine. This enables the drone's navigation system to make path planning decisions based on it. The parameterisation we have chosen is based on the following values:

\begin{itemize}
	\item The 2D location of the turbine's base on the $x$$y$ plane: $\turbinecentre \in \mathbb{R}^2$
	\item The height of the turbine's tower: $\turbineheight$ 
	\item The heading of the turbine relative to the drone's coordinate system: $\turbineheading$
	\item The length of the turbine nacelle which joins the top of the turbine tower with the blade centre: $\turbinenacelle$
	\item The rotation angle of the turbine blades: $\turbinebladerot$
	\item The length of the turbine blades: $\turbinebladelength$
\end{itemize}

Using the parameterisation above $\turbineparams = \{ \turbinecentre, \turbineheight, \turbineheading, \turbinenacelle, \turbinebladerot, \turbinebladelength \}$, a number of different turbine models can be instantiated. During the optimisation process we make use an instantiation based on points $\pointmodel$ and an instantiation based on lines $\linemodel$.

\subsubsection{Turbine Point Model}

\begin{figure}[t]
	\centering
	\includegraphics[width=0.6\linewidth]{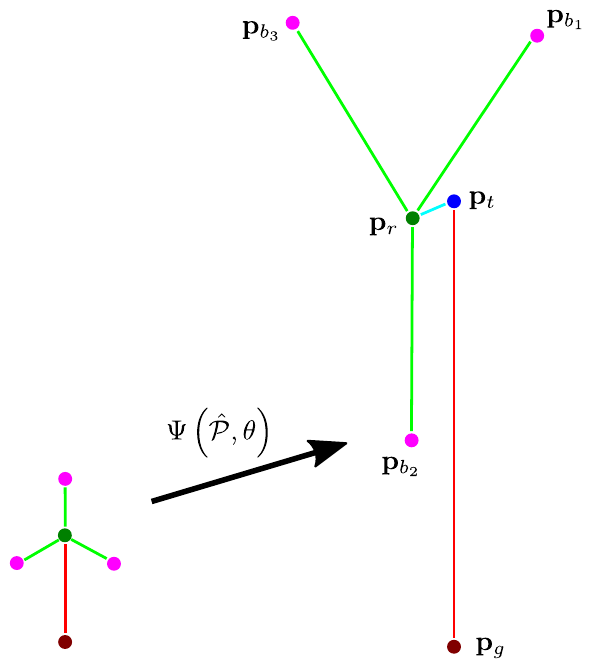}
	\caption{Image showing the model instantiation. $\hat\pointmodel$ is to the left. The instantiated model $\pointmodel$ is to the right.}
	\label{fig:model-image}
\end{figure}
The point model consists of a set of six points $\pointmodel = \{ \pointmodelpoint_g, \pointmodelpoint_t, \pointmodelpoint_r, \pointmodelpoint_{b_1}, \pointmodelpoint_{b_2}, \pointmodelpoint_{b_3}  \}$ with $\pointmodelpoint \in \mathbb{R}^3$. Each point represents a key location on the wind turbine structure. These are the turbine base $\pointmodelpoint_g$, the top of the turbine tower $\pointmodelpoint_t$, the centre of the turbine blades $\pointmodelpoint_r$, and the three tips of the blades $\pointmodelpoint_{b_i}$ where $i \in \left\{1,2,3\right\}$. To instantiate a point model $\pointmodel$ given a set of parameters $\turbineparams$, we make use of a ``default'' model $\hat\pointmodel$. This ``default'' model consists of the full set of model points $\hat\pointmodelpoint$, initialised to the following values 
\begin{equation}\label{eq:initial_point_model}
	\hat{\pointmodel} = \left\{
		\begin{array}{@{}lrc@{}}
			\hat{\pointmodelpoint}_g &  =     & \left(0,0,0\right) \\
			\hat{\pointmodelpoint}_t &  =     & \left(0,0,1\right) \\
			\hat{\pointmodelpoint}_r &  =     & \left(1,0,1\right) \\
			\hat{\pointmodelpoint}_{b_i} &  = & \left(1,0,2\right) \\
	\end{array}\right.
\end{equation} 
Note that in the ``default'' model, the tower is of unit height, the blades are of unit length, and the heading is oriented along the $x$-axis. We now define the set of functions $\Psi = \left\{ \bm\psi_g, \bm\psi_t, \bm\psi_r, \bm\psi_{b_i} \right\}$ that take the points defined in $\hat{\pointmodel}$ and the turbine parameters $\turbineparams$ to create an instance of the turbine point model model $\pointmodel$ 
\begin{equation}\label{eq:turbine_mapping}
	\pointmodel = \left\{
		\begin{array}{@{}lrl@{}}
			\pointmodelpoint_g &  =     & \bm\psi_g \,\left(\hat{\pointmodelpoint}_g, \turbineparams \right) \\
			\pointmodelpoint_t &  =     & \bm\psi_t \,\, \left(\hat{\pointmodelpoint}_t, \turbineparams \right) \\
			\pointmodelpoint_r &  =     & \bm\psi_r \, \left(\hat{\pointmodelpoint}_r, \turbineparams \right) \\
			\pointmodelpoint_{b_i} &  = & \bm\psi_{b_i}\left(\hat{\pointmodelpoint}_{b_i}, \turbineparams \right) \\
	\end{array}\right.
\end{equation}
The function $\psi_g$ that computes the location of the point at the base of the turbine $\pointmodelpoint_g$ is defined as
\begin{equation}\label{eq:compute_base_point}
	\bm\psi_g(\hat\pointmodelpoint_g, \turbineparams) = \pointmodelpoint_\turbinecentre + \hat{\pointmodelpoint}_g
\end{equation}
where $\pointmodelpoint_\turbinecentre = (\turbinecentre,0)$ maps $\turbinecentre$ into $\mathbb{R}^3$. To compute the location of the point at the top of the turbine tower $\pointmodelpoint_t$, we use
\begin{equation}\label{eq:compute_top_point}
		\bm\psi_t(\hat\pointmodelpoint_t, \turbineparams) = \pointmodelpoint_\turbinecentre + \mathbf{M}_\turbineheight \hat{\pointmodelpoint}_t \enspace,
\end{equation}
where $\mathbf{M}_\turbineheight = \mbox{diag}(1,1,\turbineheight)$ scales the input point along the $z$-axis. The point at the centre of the blades $\pointmodelpoint_r$ is computed using
\begin{equation}\label{eq:compute_hub_point}
		\bm\psi_r(\hat\pointmodelpoint_r, \turbineparams) = \pointmodelpoint_\turbinecentre + \mathbf{R}_\turbineheading \mathbf{M}_\turbinenacelle \mathbf{M}_\turbineheight \hat{\pointmodelpoint}_r \enspace,
\end{equation}
where $\mathbf{M}_\turbinenacelle =\mbox{diag}(r,1,1)$ scales the input point along the $x$-axis, and $\mathbf{R}_{\turbineheading} \in \mathbb{R}^{3x3}$ is the rotation matrix that rotates the scaled points around the $z$-axis by an angle of $\turbineheading$. The turbine blade tips $\pointmodelpoint_{b_i}$ are computed in the following manner
\begin{multline}\label{eq:compute_blade_point}
	\bm\psi_{b_i}(\hat\pointmodelpoint_{b_i}, \turbineparams) = \\ \pointmodelpoint_\turbinecentre + \mathbf{R}_\turbineheading \mathbf{M}_\turbinenacelle \left( \mathbf{R}_{\turbinebladerot_i} \mathbf{M}_\turbinebladelength \left( \hat{\pointmodelpoint}_{b_i} - \hat{\pointmodelpoint}_t \right) + \mathbf{M}_\turbineheight \hat{\pointmodelpoint}_t \right),
\end{multline}
where $\mathbf{M}_\turbinebladelength = diag(1,1,\turbinebladelength)$ scales the length of the blades along the $z$-axis. The rotation $\mathbf{R}_{\turbinebladerot_i}$ is the matrix that rotates the blades around the $x$-axis by $\turbinebladerot_i$. For the three blade tips, the values for each $\turbinebladerot_i$  are $\left(\turbinebladerot, \turbinebladerot + \frac{2}{3}\pi, \turbinebladerot + \frac{4}{3}\pi \right)$. These values produce three blade tips, rotated $120\degree$ from each other. A diagram showing an instantiated point model $\pointmodel$ can be seen in Figure~\ref{fig:model-image}.

\subsubsection{Turbine Line Model}

\noindent The second type of model we use is a line model $\linemodel = \left\{ \linemodelline_t, \linemodelline_n, \linemodelline_{b_1}, \linemodelline_{b_2}, \linemodelline_{b_3} \right\}$. These lines are defined by their end-points, which are taken from the previously defined point model $\pointmodel$
\begin{equation}\label{eq:def-line-model}
	\linemodel = \left\{
		\begin{array}{@{}lrl@{}}
			\linemodelline_t &  = & \{\pointmodelpoint_g, \pointmodelpoint_t\} \\
			\linemodelline_r &  = & \{\pointmodelpoint_t, \pointmodelpoint_r\} \\
			\linemodelline_{b_i} &  = & \{\pointmodelpoint_r, \pointmodelpoint_{b_i}\}
	\end{array} \right.
\end{equation}
The line $\linemodelline_t$ connects the base of the tower to the top of the tower. The line $\linemodelline_r$, connects the top of the tower and the centre of the blades. The lines $\linemodelline_{b_i}$ connect the centre of the blades with their tips. To instantiate an instance of $\linemodel$ given a set of parameters $\theta$, we use the set of functions $\Psi$ defined in~(\ref{eq:turbine_mapping}), applying them to each of the line end-points. This gives us a set of lines running along the centre of the main structural pieces of the wind turbine. An example of $\linemodel$ can be seen in the connecting lines in Figure~\ref{fig:model-image}. 

The reason for using a line model in combination with a point model is that during the inspection, there a numerous instances when only a small section of the wind turbine is in view. As these models are used to constrain our optimiser, if we were to rely solely on a $\pointmodel$, very few constraints would be added. By using $\linemodel$ in combination with $\pointmodel$, as long as some of the turbine structure is in view of the drone, constraints can be established.

\subsection{Turbine Projection Estimation using CNNs}\label{sec:method-projection}

\begin{figure}[t]
	\centering
	\includegraphics[width=\linewidth]{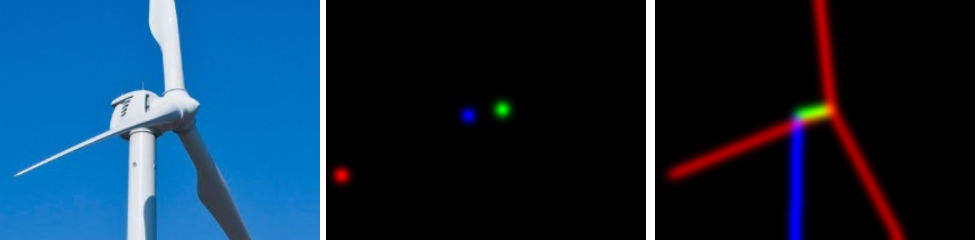}
	\caption{Example outputs from the CNN. Left) Input image. Centre) The predicted projection of $\pointmodel$, with different parts $\left\{ \projimage^{\pointmodel}_{n,g}, \projimage^{\pointmodel}_{n,t}, \projimage^{\pointmodel}_{n,r}, \projimage^{\pointmodel}_{n,b}\right\}$ coloured differently. Right) The predicted projection of $\linemodel$  with the different parts $\left\{\projimage^{\linemodel}_{n,t}, \projimage^{\linemodel}_{n,r}, \projimage^{\linemodel}_{n,b}\right\}$ coloured differently.}
	\label{fig:network-output}
\end{figure}

To enable the non-linear least squares optimisation, we need to be able to establish correspondences between model points, and their 2D locations on images of the wind turbine.
Due to the wide variation in the visual appearance of wind turbines, accurately establishing correspondences using the raw image data is difficult. To address this we use a CNN to infer the projection of our skeletal model in the images captured by the drone, i.e. the projection of $\pointmodel$ and $\linemodel$ into the camera coordinates, from the drone's current view of the turbine. This results in a set of projection images $\mathbf{P}_n = \left\{ \projimage^{\pointmodel}_{n,g}, \projimage^{\pointmodel}_{n,t}, \projimage^{\pointmodel}_{n,r}, \projimage^{\pointmodel}_{n,b}, \projimage^{\linemodel}_{n,t}, \projimage^{\linemodel}_{n,r}, \projimage^{\linemodel}_{n,b
}\right\}$. Important to note are the superscripts referring to the type of model, the subscript $n$ associating the images with a vertex in the pose graph (see Section~\ref{sec:method-optimisation}) and the second subscript referring to the type of structure within the model. The set of projection images are used to establish correspondences between the models, and locations in the images. Example outputs of the CNN can be seen in Figure~\ref{fig:network-output}. The network is trained using manually labelled images of wind turbines. Readers are referred to~\cite{Moolan-Feroze2019} for a full description of the CNN and training process.

\subsection{Optimisation}\label{sec:method-optimisation}

The parameter estimation problem is solved using a non-linear least squares optimisation over a pose graph $\mathcal{G} = \langle \mathcal{V}, \mathcal{E}\rangle$ and a set of turbine parameters $\theta$. The graph vertices $\mathcal{V} = \left\{ \vertex_n \right\}_{n=1}^{N}$, where $N$ is the total number of vertices in the graph, each contain a full 6DoF pose $\mathbf{T}_n$, with the orientation represented by a quaternion $\mathbf{q}_n$ and a translation $\mathbf{t}_n$. The edges connect the vertices sequentially $\mathcal{E} = \left\{ i,i+1 \right\}_{i=1}^{N-1}$, mirroring the sequential addition of new vertices to the graph during inspection. The total set of parameters that we are optimising is then $G = \left\{\theta\right\} \bigcup \left\{ \mathbf{q}_n, \mathbf{t}_n \right\}_{n=1}^{N}$. We define the cost function to be optimised as  
\begin{equation}\label{eq:cost-function}
	E_{\text{cost}} = 
		\lambda_{\pointmodel} E^{\pointmodel}_{\text{unary}} + 
		\lambda_{\linemodel} E^{\linemodel}_{\text{unary}} + 
		\lambda E_{\text{pairwise}} \enspace.
\end{equation}
The two unary terms measure the accuracy of the current parameter estimates based on the image data using points from $\pointmodel$ and points from $\linemodel$. The pairwise term regularises the cost function by penalising poses that differ from pose measurements obtained by the GPS/IMU. 

\subsubsection{Unary Cost Function Term}

Both unary terms $E_{\text{unary}}$ in~(\ref{eq:cost-function}) are represented by the same function. We use the superscript to designate that one is computed using the points from $\pointmodel$ and one computed using the points from $\linemodel$. We will use the notation $\model$ to equal either $\pointmodel$ or $\linemodel$. In this section we present two different forms of $E^{\model}_{\text{unary}}$; one based on 2D point correspondences following the work in~\cite{Moolan-Feroze2019} $\pointterm$, and a new method $\imageterm$ based on direct evaluation of the pixel values in the projection images $\mathbf{P}_n$. Necessary for both these formulations of the unary term is the following projection function
\begin{equation}\label{eq:projection}
	\mathbf{a}_{n,m} = \Pi\left(\vertex_n, \pointmodelpoint_m\right) = \cammat [ \mathbf{R}\left(\mathbf{q}_n\right) | \mathbf{t}_n ] \enspace \pointmodelpoint_m \enspace.
\end{equation}
Typical in monocular tracking applications, this function allows us to transform a point $\pointmodelpoint_m$ from the world frame into camera image coordinates $\mathbf{a}_{n,m}$ using the transformation $\mathbf{q}_n$ and $\mathbf{t}_n$ held in a vertex $\vertex_n$, and $\mathbf{K}$ which represents the camera intrinsics. The first version of the unary term we present is based on 2D point correspondences and is formulated as
\begin{equation}\label{eq:point-data-term}
	\pointterm =  
	\sum\limits_{\vertex_n \in \mathcal{V}}
	\sum\limits_{\hat\pointmodelpoint_m \in \hat\model} \| \Pi\left( \vertex_n, \bm\psi_m (\hat\pointmodelpoint_m, \theta) \right) - \mathbf{a}'_{n,m} \|^2 \enspace.
\end{equation}
The value $\mathbf{a}'_{n,m}$ is the 2D correspondence to the model point $\hat\pointmodelpoint$, on the projection image $\projimage^{\model}_{n,m}$. This term transforms the model point into the world frame using the functions $\Psi$, and then projects it into the image using~(\ref{eq:projection}). One important process is establishing these correspondences prior to optimising. Depending on the type of model, correspondences are found in different ways. 
\begin{figure}[t]
	\centering
	\includegraphics[width=1.0\linewidth]{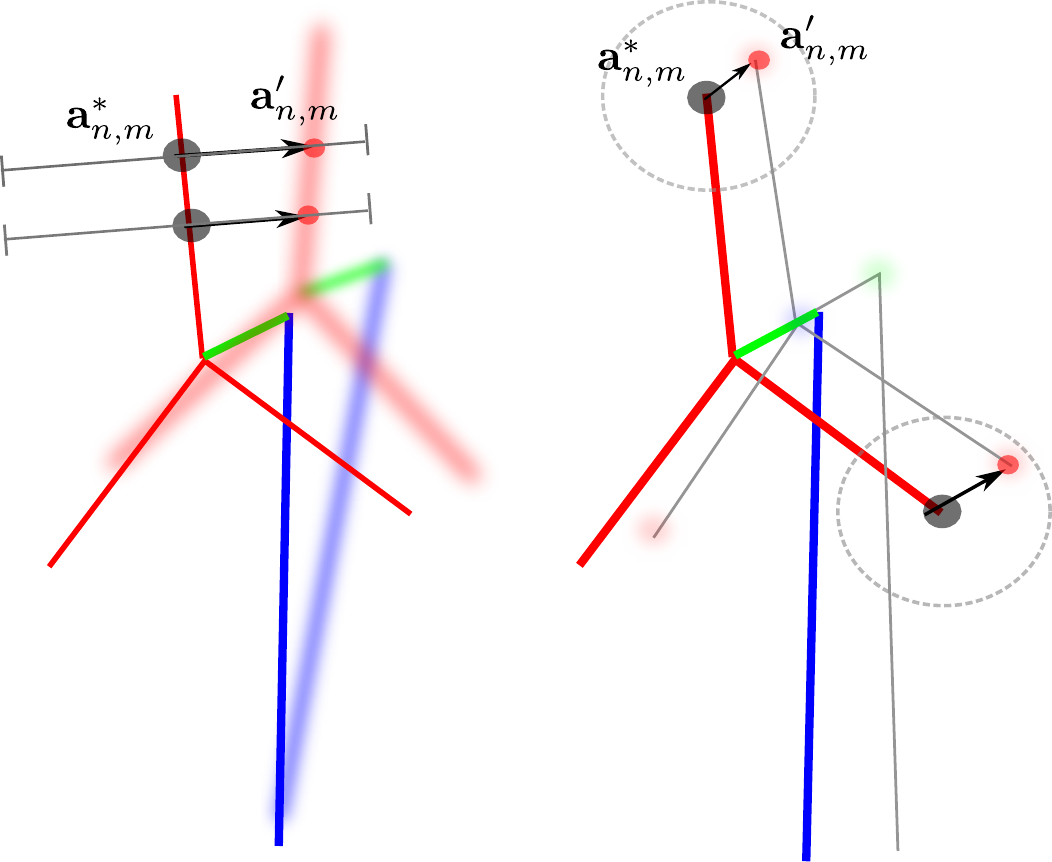}
	\caption{Figure showing how correspondences are established. Model projections $\projimage$ inferred by the CNN are shown as blurred points and lines and the model projection of $\linemodel$ and $\pointmodel$ using the current pose estimate is shown in bold colours. Left) Line-based correspondences are found using a perpendicular line search.  Right) Point based correspondences are found by searching within a circular window.}
	\vspace{-1em}
	\label{fig:constraint_images}
\end{figure}

The first uses the point model $\pointmodel$. We take each of the current estimates of the poses in the graph $\vertex_n^* \in \mathcal{V}^*$ and the current estimate of the parameters $\theta^*$ and project each of the model points $\hat\pointmodelpoint_m$ into the image. 
\begin{equation}\label{eq:point-correspondence}
	\mathbf{a}^*_{n,m} = \Pi\left( \vertex_n^*, \bm\psi_m (\hat\pointmodelpoint_m, \theta^*) \right) \enspace.
\end{equation}
We then search the image pixels in $\projimage^{\pointmodel}_{n,m}$ around $\mathbf{a}^*_{n,m}$ and find the pixel with the largest value. This pixel, if the value is above a threshold is selected as $\mathbf{a}'_{n,m}$. This process is illustrated in Figure~\ref{fig:constraint_images}.

The second method using the line model $\linemodel$ is somewhat more involved. The reason for using the line model in the optimiser is that as the drone is often close to the turbine during the inspection, the points in $\pointmodel$ are not always visible. Using $\linemodel$ allows us to establish correspondences in-between the line end points. To do this we split each of the lines in $\linemodel$ into points $\hat\pointmodelpoint_m$, where $m \in \{t,r,b\}$ and use these to establish correspondences. 

Similar to the points in $\pointmodel$, for each split point $\hat\pointmodelpoint_m$, we find the 2D projection $\mathbf{a}^*_{m,n}$ using~(\ref{eq:point-correspondence}). Using the projected location, we then perform a perpendicular line search from $\mathbf{a}^*_{n,m}$. The pixel along the line in $\projimage^{\linemodel}_{n,m}$ with the highest value above a threshold is then chosen as $\mathbf{a}'_{n,m}$. This process is illustrated in Figure~\ref{fig:constraint_images}. For a more detail on the process of establishing correspondences, please see~\cite{Moolan-Feroze2019}.

The second formulation of the unary cost function $\imageterm$ is based on directly sampling the pixel values in $\projimage^{\model}_{n,m}$ using the projection of the model points into the image and interpolating the values during optimisation.  The output of the CNN can be viewed as a heat map, with pixel values ranging between $[0,1]$. The higher the pixel value, the great the expectation that the corresponding turbine feature lies at that location. We take this as a probability, and optimising the cost function becomes a process of finding the parameters that project the point model to the locations on the image with the highest probability, corresponding to the minimum of the sum of the negative log over all points, i.e.
\begin{equation}\label{eq:image-data-term}
	\imageterm =
	\sum\limits_{\vertex_n \in \mathcal{V}}
	\sum\limits_{\hat\pointmodelpoint_m \in \hat\model} - \log \left( \Gamma \left( \mathbf{P}^{\mathcal{M}}_{n,m}, \Pi\left(  \vertex_n, \Psi_m (\hat\pointmodelpoint_m, \theta) \right) \right) \right) \enspace.
\end{equation}
The function $\Gamma$ uses bicubic interpolation to evaluate the pixel value under the projected point using Cubic Hermite Splines. Importantly, this type of interpolation is differentiable, allowing us to integrate the function into the non-linear optimiser. 

In this application, there are a number of benefits of using the direct image interpolation method $\imageterm$ over the 2D correspondence method $\pointterm$. It means we don't need to establish correspondences prior to optimisation, reducing the complexity of the process. More importantly, the interpolation method allows the model point projections to move freely in image space during optimisation. This is especially important for the points from $\linemodel$. If we use the 2D point correspondences as defined in $\pointterm$, the optimiser is discouraged from using update steps that will move line points in the same direction as the corresponding line in the projection image $\projimage$. This is unwanted, as the cost incurred by a line point should be equal anywhere along the length of the corresponding line in $\projimage$. 

\subsubsection{Pairwise Cost Function Term}

The second term of the cost function $E_{\text{pairwise}}$ is used to constrain the pose parameters during optimisation such that they don't differ too far from the original estimates of the drone's pose measured using GPS/IMU. To enable this we compute the relative poses  $\mathbf{T}_{i,j} \,\, \forall \,\, \{i,j\} \in \mathcal{E}$ in the following way
\begin{equation}\label{eq:relative-pose-t-offset}
	\mathbf{T}_{i,j} =
	\begin{bmatrix}
		\mathbf{t}_{i,j}\\[0.5em]
		\mathbf{q}_{i,j}
	\end{bmatrix}
	=
	\begin{bmatrix}
		\mathbf{R}\left(\mathbf{q}_i\right)^T \left(\mathbf{t}_{j} - \mathbf{t}_i \right) \\[0.5em]
		\mathbf{q}^{-1}_i * \mathbf{q}_{j}
	\end{bmatrix} \enspace,
\end{equation}
where $\mathbf{R}\left(\mathbf{q}_i\right)$ is the rotation matrix formed from the quaternion rotation $\mathbf{q}_i$. Using relative measurements as constraints rather than absolute measurements is beneficial in that it allows the optimised poses to drift over time -- correcting for drift in the GPS/IMU --  while harshly penalising abrupt differences between successive key frames. The full pairwise term is then defined as
\begin{equation}\label{eq:relative-pose-diff}
	E_{\text{pairwise}} = \sum\limits_{i,j \in \mathcal{E}} \enspace\enspace  \mathbf{C}
	\begin{bmatrix}
		\hat{\mathbf{t}}_{i,j} - \mathbf{t}_{i,j} \\[0.5em]
		2 \times \text{Vec} \left(  \hat{\mathbf{q}}_{i,j} * \mathbf{q}_{i,j}^{-1} \right)
	\end{bmatrix} \enspace,
\end{equation}
where `Vec' corresponds to the vector component of the quaternion. The matrix $\mathbf{C}$ is an information matrix, which encodes the certainty about the relative pose measurements obtained by the GPS/IMU. Due to limitations on our system, we don't have access to this data during the flight. Instead we fix this matrix using a set of reasonable values. Further details can be found in \cite{Moolan-Feroze2019}.

\vspace{2mm}
\noindent To optimise~(\ref{eq:cost-function}), we use the Levenberg-Marquardt~\cite{Marquardt2005} non-linear least squares method. This gives us our fitted pose and turbine parameter values. We make use of the Ceres Solver~\cite{ceres-solver} software library to perform the optimisation process.

%
\section{EVALUATION}\label{sec:evaluation}

To evaluate our work we present two types of experiments. The first in Section~\ref{sec:evaluation-simulated} relies on simulated data, allowing us to empirically evaluate the performance of our method against a known ground truth. The second set in Section~\ref{sec:evaluation-real} applies our work to real wind turbine inspection data. As this data does not have a known ground truth, we only provide a qualitative evaluation of the results.

\subsection{Simulated Data}\label{sec:evaluation-simulated}
\begin{figure}[t]
	\centering
	\includegraphics[width=0.95\linewidth]{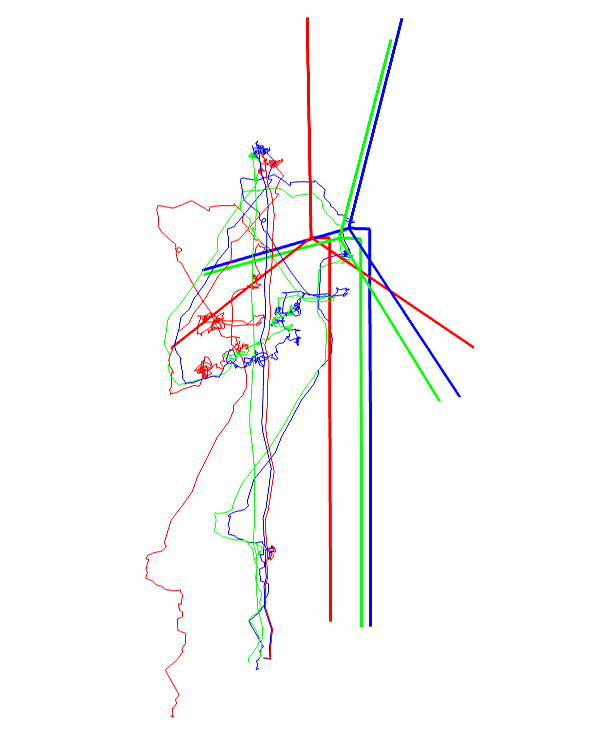}
	\caption{Visual example of the combined pose and model optimisation process. The green turbine corresponds to the ground truth $\mathcal{P}$, the red to the initial configuration of $\mathcal{P}$ and the blue to the optimised configuration of $\mathcal{P}$. Similarly, the green path corresponds to the ground truth positions of the drone, the red to the initial positions and the blue to the optimised positions.}
	\label{fig:synth_output_figure}
\end{figure}
\begin{figure}[t]
	\centering
	\includegraphics[width=\linewidth]{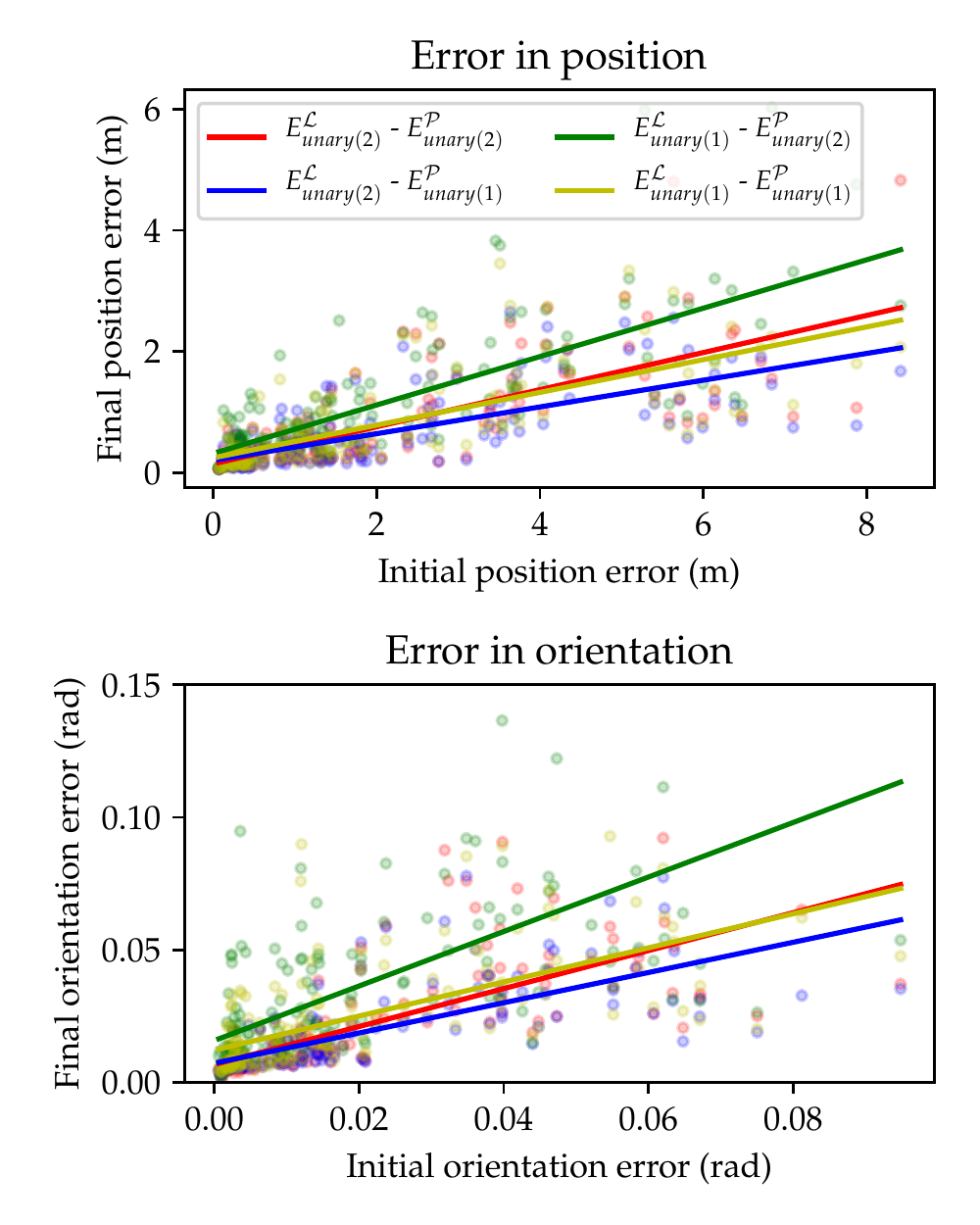}
	\caption{Plots comparing the performance of the different types of unary cost on location accuracy (top) and orientation accuracy (bottom). The plot legend labels correspond to the cost term used for the points in $\mathcal{L}$ and $\mathcal{P}$. The solid lines show the LS fit line for each configuration.}
	\label{fig:constraint_figure}
\end{figure}

To provide a qualitative evaluation of our method we applied it to a set of simulated inspection data. To generate the synthetic data, we extracted pose graphs and turbine parameter sets from a number of real inspection flights which we use as pseudo ground truths. This provides us with a representative set of data for evaluation. For each of these ground truths we applied varying levels of noise to the data. For the pose graphs, we sequentially added noise to the position and orientations, allowing us to simulate the sort of drift one might expect from using GPS/IMU readings during a flight. For the turbine parameters we added error from a Gaussian distribution to each of the values. Due to the different scales for the parameters in $\theta$, the $\sigma$ values for each parameter were different. Using this data we evaluated a number of different aspects of our method. An example of the output can be seen in Figure~\ref{fig:synth_output_figure}.

\subsubsection{Unary cost function comparison}

The first aspect we evaluated was the difference in performance of the two unary cost terms we defined in Section~\ref{sec:method-optimisation}. These are the one based on 2D point correspondences $\pointterm$ and the one based on direct image interpolation $\imageterm$. As we have two different sets of model points that these can be applied to -- the points from $\pointmodel$ and the points from $\linemodel$ -- we compared a total of 4 different configurations. We applied each of these configurations to our set of synthetic data and recorded the mean error in position and orientation prior to optimisation, and the same values after optimisation. The results of the experiments can be seen in Figure~\ref{fig:constraint_figure}. As the plots show, the method that uses $\imageterm$ for $\linemodel$ and the $\pointterm$ for $\pointmodel$ performs best for both position and orientation. The method that performs the worse is when this configuration is reversed. This performance is expected due to the factors highlighted in Section~\ref{sec:method-optimisation}. As the points in $\pointmodel$ correspond to actual 2D correspondences in the image, using the point correspondence cost $\pointterm$ makes sense. As the points in $\linemodel$ should contribute a similar amount to the cost at any location along the line of the corresponding turbine feature, the more relaxed $\imageterm$ function is appropriate. Given these results, for the remainder of the experiments we use $\imageterm$ for $\linemodel$ and $\pointterm$ for $\pointmodel$.

\subsubsection{Combined Optimisation Comparison} 

\begin{figure}[t]
	\centering
	\includegraphics[width=\linewidth]{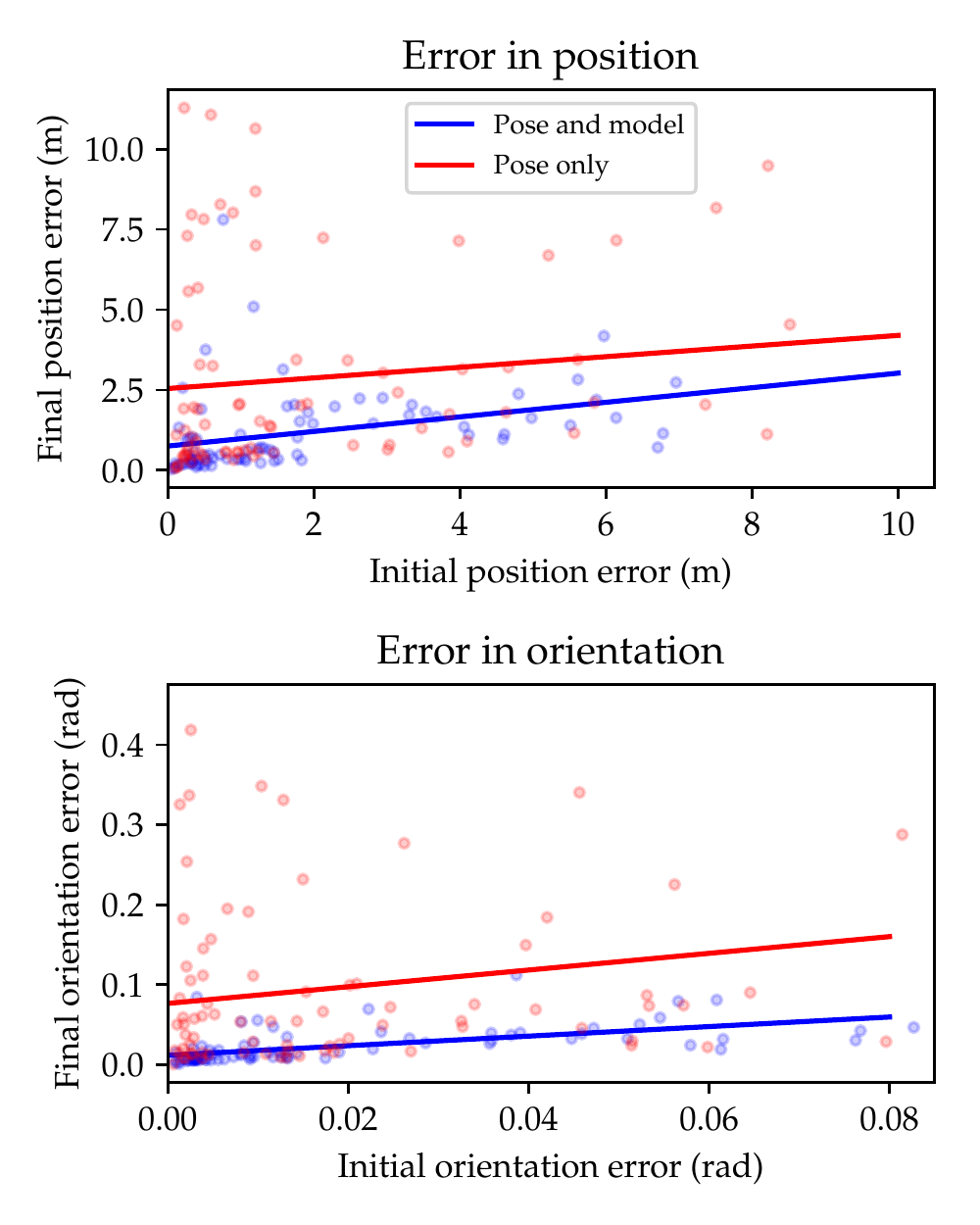}
	\caption{Plot comparing the performance of our combined method over the method that optimises only the pose. Top) Position error, Bottom) Orientation Error.}
	\label{fig:compare_figure}
\end{figure}

\begin{figure}[t]
	\centering
	\includegraphics[width=\linewidth]{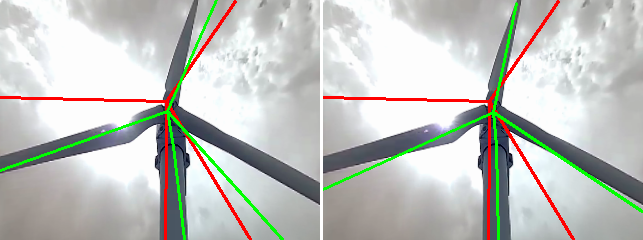}
	\caption{Image comparing pose only optimisation with joint pose and model optimisation on real data. The red turbine projection is the initial estimate. The green projection is the optimised estimate. Left) Output from pose only optimisation. Right) Output from combined optimisation.}
	\label{fig:compare_real}
\end{figure}

The second aspect of our method we evaluated was the benefit of simultaneously optimising both the pose and turbine parameters over optimising just the pose. To do this we applied our method to the set of synthetic data twice; once where both sets of parameters were optimised, once where only the pose was optimised. We then evaluated the error of both methods. The results of this experiment can be seen in Figure~\ref{fig:compare_figure}. As these figures show, when we are optimising both sets of parameters, the method is significantly improved over optimising the pose alone. The reason for this is that errors in the turbine parameters will propagate into errors in the pose of the drone, if the model parameters are fixed during optimisation. The most obvious instance of this is when the value $\phi$ -- which sets the rotation of the blades -- is wrong. During optimisation, this will propagate into error in the roll as the drone seeks to align the incorrect model with the image data. We can see this in Figure~\ref{fig:compare_real}.

\subsubsection{Recovering the Turbine Parameters}

\begin{figure}[t]
	\centering
	\includegraphics[width=\linewidth]{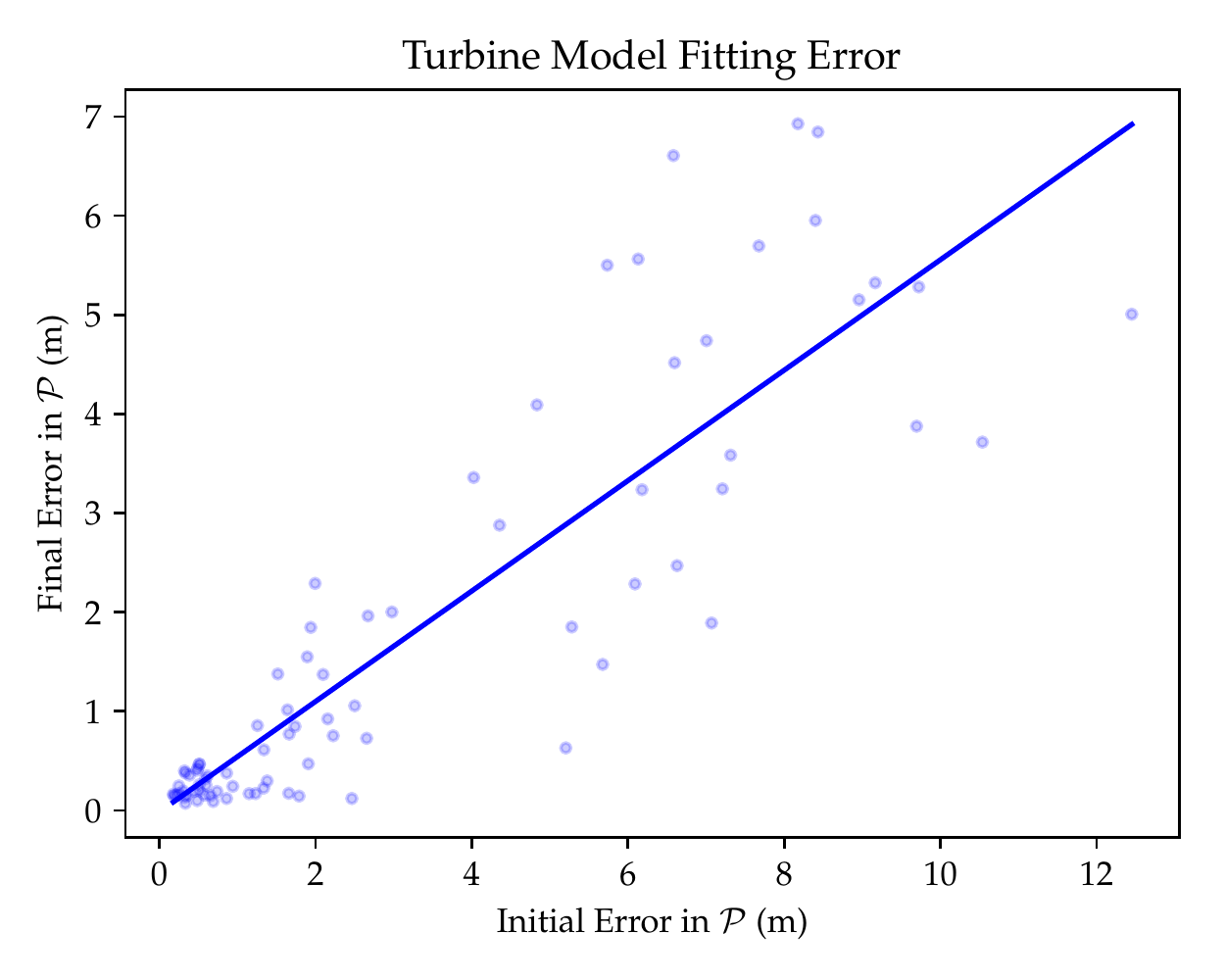}
	\caption{Plot showing the performance of our method for fitting the turbine model parameters. Plot shows initial error in meters against error after optimisation.}
	\label{fig:turbine_figure}
\end{figure}

The final set of synthetic evaluations we performed was to establish the ability of the method to optimise the parameters of the turbine model. To do this we applied the set of functions $\Psi$ to the ground truth parameters, the initial parameters (those with noise added) and the optimised parameters. We then computed the mean distance error across all the points in $\pointmodel$ for the initial model and the optimised model and plotted them against each other. The results of this can be seen in Figure~\ref{fig:turbine_figure}. We can see that as expected, when there is more error in the initial turbine parameters, the optimiser is less able to recover to the true pose. We do see however, that the method is able to recover around half the error introduced into the initial parameters.

\subsection{Real Data}\label{sec:evaluation-real}
\begin{figure*}[t]
	\centering
	\includegraphics[width=\linewidth]{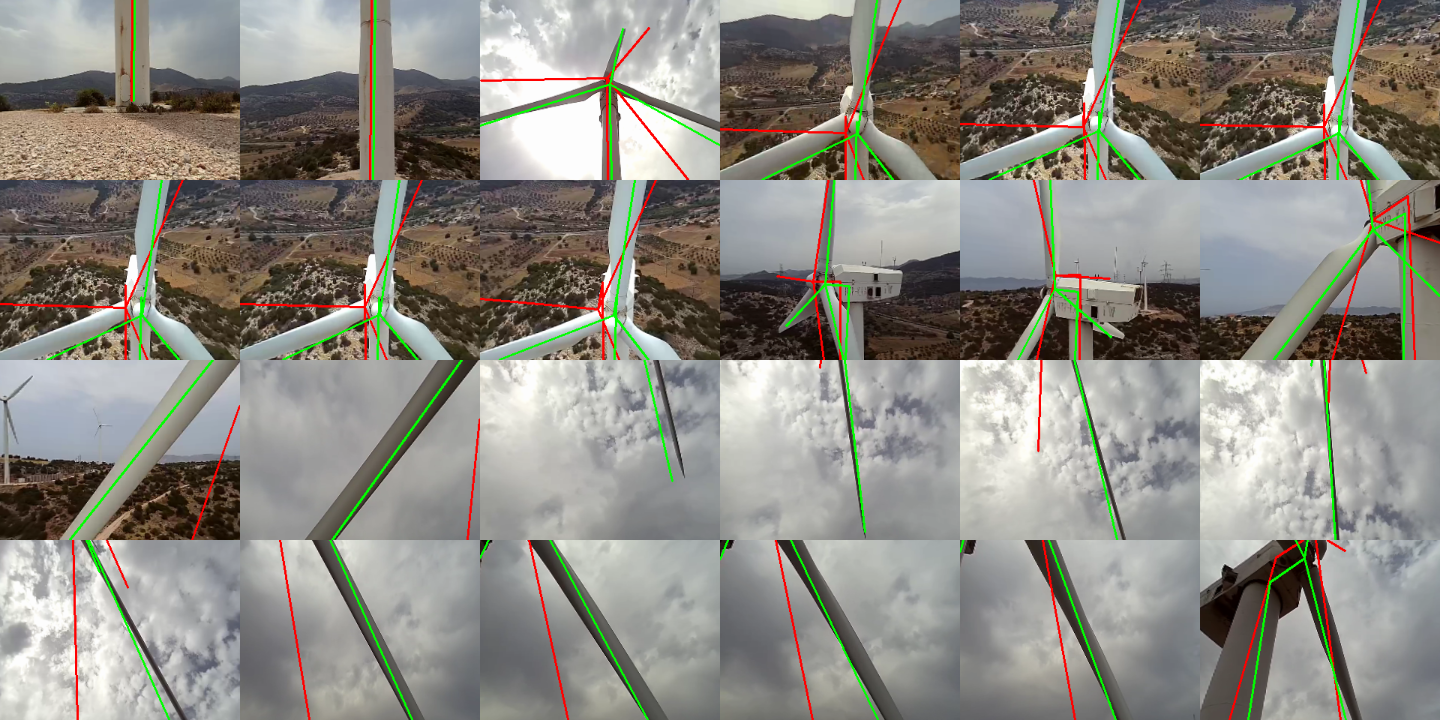}
	\caption{Frames taken from a real inspection. The red turbine is the reprojection using the initial pose parameters and turbine model parameters. The green turbine is the reprojection from the optimised pose parameters and turbine model parameters.}
	\label{fig:full_flight}
\end{figure*}

To give an example of the performance of our method we applied it to some recordings of real inspection flights. For this data, the initial pose estimates are measured using GPS/IMU readings and the turbine model parameters were initialised by eye. In Figure~\ref{fig:compare_real}, we give an example of the benefit of optimising both model and pose parameters simultaneously. As we can see, when there is error in blade rotation value $\phi$, if we are not able to correct for that error during the flight, the optimiser will be encouraged to correct for it by incorrectly adding roll to the orientation.

In Figure~\ref{fig:full_flight}, we show a series of images showing the reprojection of the turbine model. We can see that our method is able to correct for a large proportion of the error in the initial estimate of the turbine and pose.

%
\section{CONCLUSIONS}\label{sec:conclusions}

In this work we have presented a method for the simultaneous optimisation of a drone's pose as well the parameters of a wind turbine model during an inspection flight. We present a set of functions that allow us to integrate the model parameter fitting into a non-linear least squares optimiser. In addition we also present a new cost function term which is better suited to the image data than the traditional corresponding points method. In our evaluations, we have compared the two types of unary cost terms, shown the benefit of the simultaneous optimisation using both simulated and real data, and measured how effective our method is at correcting errors in the model parameters. 

There are a number of avenues open for future work. One of these is the incorporation of extra sensor measurements into the optimisation. This could take the form of depth data from a LiDAR or from stereo cameras. We also aim to properly integrate the GPS/IMU measurements using a visual inertial odometry method.

\section*{ACKNOWLEDGEMENTS}

The authors acknowledge the support of Innovate UK (project number 104067). The work described herein is the subject of UK patent applications GB1815864.2 and GB1902475.1.


\bibliographystyle{IEEEtran}
\bibliography{mybib}

\end{document}